\documentclass[10pt,twocolumn,letterpaper]{article}

\usepackage{cvpr}              %

\usepackage[dvipsnames]{xcolor}

 \usepackage{graphicx}
\usepackage{amsmath}
\usepackage{amssymb}
\usepackage{booktabs}
\usepackage{multirow}
\usepackage{multicol}
\usepackage{float}

\newcommand{\mypar}[1]{\vspace{0.5mm}\noindent\textbf{#1}}
\newcommand{\methodname}{AffordanceLLM}

\definecolor{cvprblue}{rgb}{0.21,0.49,0.74}
\usepackage[pagebackref,breaklinks,colorlinks,citecolor=cvprblue]{hyperref}

\def\paperID{3} %
\def\confName{CVPR}
\def\confYear{2024}

\title{\methodname: Grounding Affordance from Vision Language Models}

\author{
Shengyi Qian\thanks{The work was done during an Amazon internship.} \hspace{5mm} Weifeng Chen \hspace{5mm} Min Bai \hspace{5mm} Xiong Zhou \hspace{5mm} Zhuowen Tu \hspace{5mm} Li Erran Li\\
  AWS AI, Amazon\\
 {\small \url{https://jasonqsy.github.io/AffordanceLLM}}\\
}

\makeatletter
\g@addto@macro\@maketitle{
\vspace{-3.3em}
\begin{figure}[H]
   \setlength{\linewidth}{\textwidth}
\setlength{\hsize}{\textwidth}
\centering
\includegraphics[width=\linewidth]{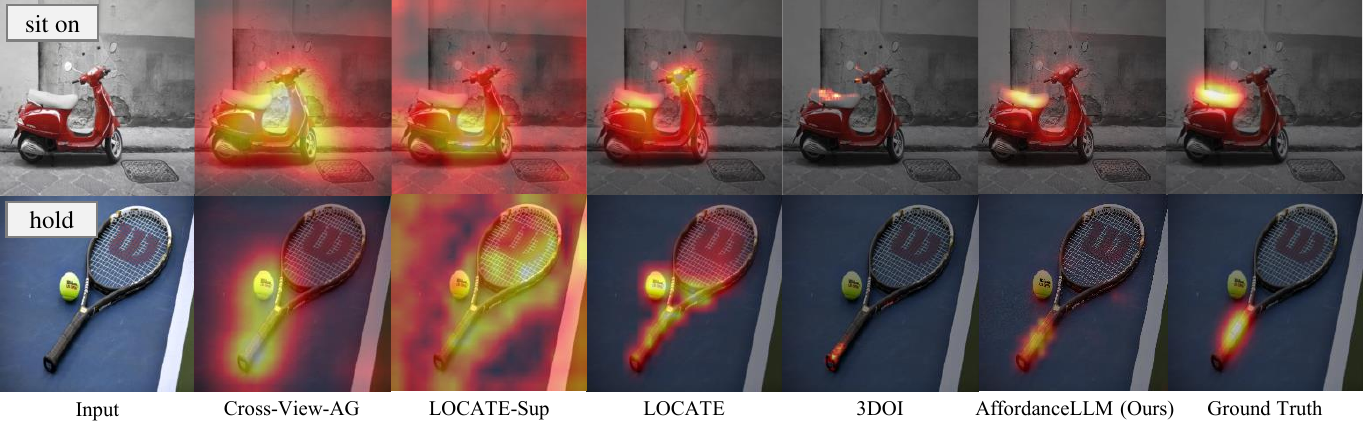}
\vspace{-2em}
\caption{{\bf Illustration for the affordance grounding task}. The input is a single image and the corresponding action (e.g, ``hold''). The output is a heatmap which highlights regions one can interact. 
We aim to enhance the generalization capability of affordance grounding to in-the-wild objects that are unseen during training, by developing a new approach, \methodname, that takes the advantage of the rich knowledge from large-scale vision language models~\cite{liu2023llava} beyond the supervision from the training images.
}
\label{fig:intro}
\end{figure}
}
\makeatother

\begin{document}
\maketitle
\begin{abstract}
Affordance grounding refers to the task of finding the area of an object with which one can interact. It is a fundamental but challenging task, as a successful solution requires the comprehensive understanding of a scene in multiple aspects including detection, localization, and recognition of objects with their parts, of geo-spatial configuration/layout of the scene, of 3D shapes and physics, as well as of the functionality and potential interaction of the objects and humans. 
Much of the knowledge is hidden and beyond the image content with the supervised labels from a limited training set. 
In this paper, we make an attempt to improve the generalization capability of the current affordance grounding by taking the advantage of the rich world, abstract, and human-object-interaction knowledge from pretrained large-scale vision language models~\cite{liu2023llava}. 
Under the AGD20K benchmark, our proposed model demonstrates a significant performance gain over the competing methods for in-the-wild object affordance grounding. 
We further demonstrate it can ground affordance for objects from random Internet images, even if both objects and actions are unseen during training.

\end{abstract}

\section{Introduction}
\label{sec:intro}

Grounding affordance from a single image is a fundamental problem in computer vision. It forms the stepping stone to downstream tasks such as understanding human-object interaction~\cite{Chao15,Gkioxari18,Shan20}, visual navigation~\cite{Kumar18}, and object manipulation~\cite{bahl2022human,hsu2023ditto}. 
Past approaches generally use human demonstrations as supervision to advance this field with tremendous success~\cite{nagarajan2019grounded,luo2022learning,luo2022grounded,li2023locate}. While such approaches perform well on objects and actions seen during training, they struggle when generalizing in the wild, i.e. on novel objects unseen during training (Fig.~\ref{fig:intro}).

The difficulties in generalization arise from the fact that affordance grounding is a challenging task that requires comprehensive understanding of an image from multiple aspects. 
A successful solution requires an understanding of the 3D geometry and functionality of objects and parts, of the actions and intentions of the executing agent,  
of the potential interaction between object and human, as well as of the spatial configuration of the environment. 
Much of these knowledge lies beyond the ground-truth localization/recognition of objects and parts provided as heatmaps in a limited training set.

\begin{figure}
    \centering
    \includegraphics[width=\linewidth]{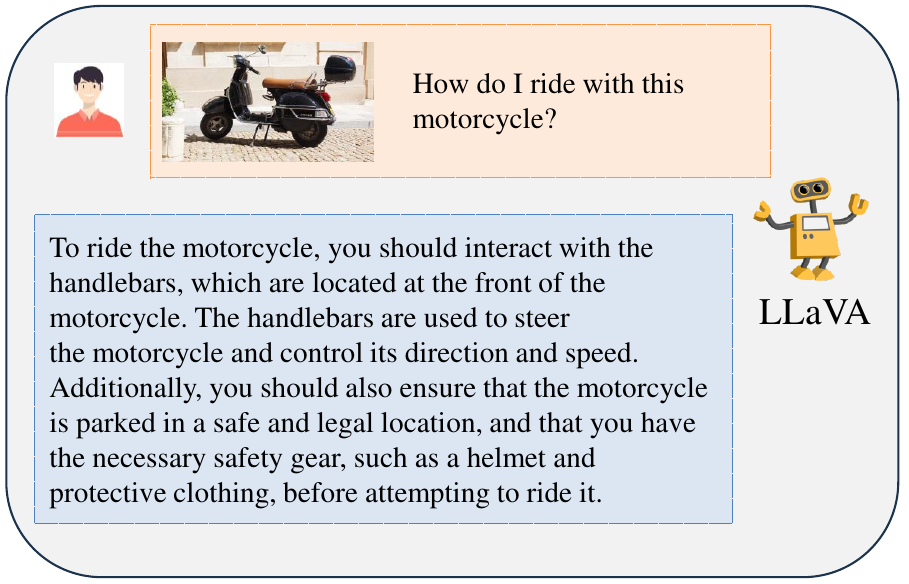}
    \vspace{-2em}
    \caption{State-of-the-art vision language models, such as LLaVA~\cite{liu2023llava}, has rich human-object-interaction knowledge, thanks to the large-scale text pretraining.
    Given a question about how to interact with an object, it typically gives a reasonable solution.}
    \vspace{-1em}
    \label{fig:llava}
\end{figure}

In this paper we make attempts to improve affordance grounding in the wild by leveraging the rich world, abstract, and human-object-interaction knowledge embedded in large-scale Vision Language Models (VLMs). 
With large-scale text pretraining, modern VLMs such as GPT-4~\cite{openai2023gpt4}, LLaVA~\cite{liu2023llava} and Blip-2~\cite{li2023blip2} have a rich reservoir of world knowledge, as demonstrated by their extraordinary capabilities in answering visually grounded common sense questions~\cite{bubeck2023sparks}. World knowledge is instrumental to affordance reasoning --- when presented with an image of a motorcycle and questioned about ``How do I ride with this motorcycle?'', LLaVA answers ``To ride the motorcycle, you should interact with the \underline{handlebars}...'' (Fig.~\ref{fig:llava}), which exhibits commonsensical understanding of affordance. Affordance models equipped with similar world knowledge have a better chance generalizing to the wild than a model that purely learns from limited affordance demonstration data.

Beside world knowledge, another novel factor we introduce to improve affordance reasoning is 3D geometry, as it holds rich information of object functionality.  A cylindrical area, for example a handle or a stick, is closely related to the action of grabbing or holding, regardless of its color or texture. Similarly, a flat part, for example the surface of a chair or a bench, might indicate areas suitable for sitting or lying. Relating 3D geometries to actions allows us to bypass the difficulties in handling variations in visual appearances, and thus facilitates generalization. 

We propose a novel approach, \textit{\methodname}, that reflects the above intuitions.
Our approach builds upon a VLM backbone (LLaVA~\cite{liu2023llava}) to tap into the its world knowledge.
We achieve it by extending the backbone with a mask decoder and a special token \texttt{<mask\_token>}, which are used to predict an affordance map.
The whole model can be trained end-to-end.
Additionally, we introduce depth maps as 3D information in parallel to RGB images as input to our network, 
with the goal of eliciting geometric reasoning capability from the network. We found both designs significantly improve performance.

We evaluate our method on the AGD20K~\cite{luo2022learning} benchmark, as this is the only large-scale affordance grounding dataset with accurate action and object labels.
We re-split the benchmark to test capability of models to generalize to object categories unseen during training.
We show that our approach outperforms all state-of-the-art baselines by a large margin.
We take a further step to validate the generalization ability by testing our approach on random Internet images. It produces reasonable affordance maps on object categories very different from the ones in training set. Moreover, it even possesses some capability of generalizing to completely novel actions.

In summary, our contributions are as follows:
\begin{enumerate}
\item We introduce the first-ever affordance grounding approach that leverages the rich world knowledge embedded in pretrained VLMs, enabling the model to generalize beyond training data;
\item We demonstrate the importance of 3D information in affordance grounding;
\item Our proposed approach generalizes to novel objects and outperforms all state-of-the-art approaches on AGD20K. It even shows evidence that it could generalize to novel actions.
\end{enumerate}

\section{Related Work}

\begin{figure*}[t]
    \centering
    \includegraphics[width=\linewidth]{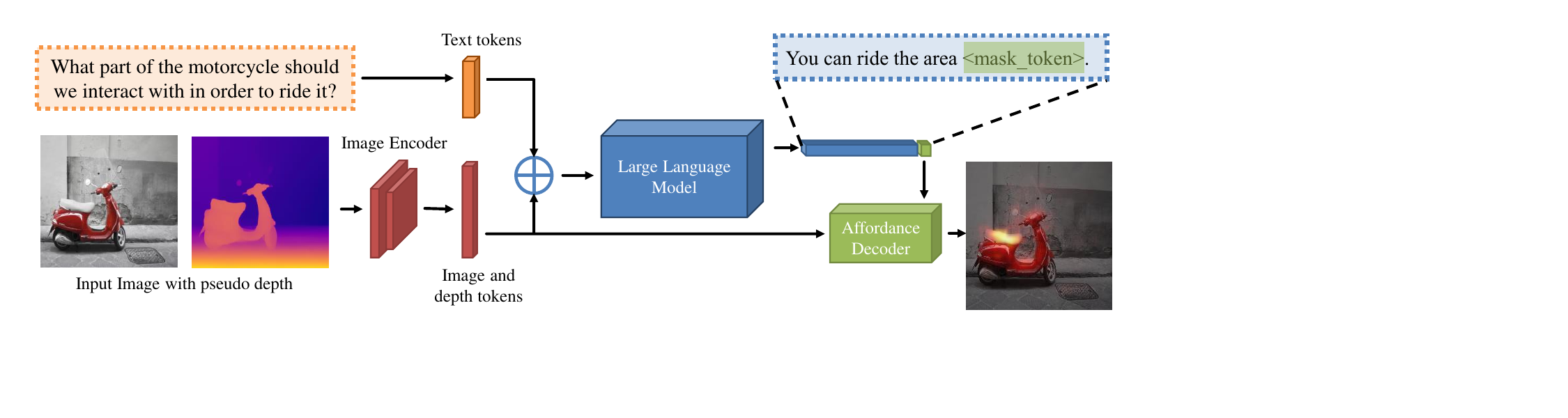}
    \vspace{-2em}
    \caption{\textbf{Overview of \methodname.} The inputs of our model includes a single image and a text prompt related to interaction. We use OWL-ViT~\cite{minderer2022simple} as the image encoder to generate image features and project it into the same hidden dimension as the large language model. As well, we use a tokenizer to encode the text prompt. The text features and image features are concatenated together and feed into the LLM. The LLM is fine-tuned to predict a special token, which is used as a query to the mask decoder to generate the final affordance map.}
    \label{fig:approach}
    \vspace{-1em}
\end{figure*}

\mypar{Eliciting World Knowledge from Vision Language Models.} 
Foundational Vision Language Models that bridge images and language have a rich reservoir of world knowledge, and recent researches have been tapping into it to make advancement in vision tasks. 
The joint visual-language embedding space learnt from simple image-text pair~\cite{radford2021clip,desai2021virtex} has made it possible to improve open-world detection~\cite{minderer2022simple,liu2023grounding, ma2023codet,Shi_2023_ICCV}, and segmentation~\cite{wang2022ofa,luddecke2022clipseg,li2022lseg,xu2023masqclip}. The world knowledge here is the correspondence between visual and language concepts.

Large language models (LLMs) trained on trillions of tokens contain even richer world knowledge and are capable of answering common-sense questions. Coupled with vision inputs, the resulting multi-modal LLMs are brought in to solve complex vision problems. 
For example, Kosmos-2 and Groundhog incorporate the reasoning skill of LLMs to generate bounding boxes and segmentation masks~\cite{peng2023kosmos2,zhang2024groundhog,lai2023lisa}.
3D LLM further extend LLMs to reason about 3D scenes, including visual grounding and navigation~\cite{yang2023llm,zhou2023navgpt,dai2023think,hong20233dllm,huang2023embodied,li20243dmit,huang2023chat}. 
For robotics, PaLM-E and RT2 transfer the knowledge from visual-language domains into motion planing and manipulation~\cite{driess2023palm,brohan2023rt2}.
Our approach embodies the same idea to transfer the world knowledge from VLMs, but applies it on a novel setting -- solving visual affordance grounding.

\mypar{Affordance Grounding.} 
Understanding object affordance from a single image is an important
step towards embodied visual intelligence, and researchers have built many different approaches to endow machines to have this ability.
Nagarajan \etal first proposes to ground object affordance from Internet videos~\cite{nagarajan2019grounded}.
Fang \etal constructs an object affordance dataset based on product review videos~\cite{fang2018demo2vec}.
Luo \etal annotates the first large-scale affordance dataset and call it AGD20K~\cite{luo2022learning}.
LOCATE~\cite{li2023locate} is the state-of-the-art approach on AGD20K.
More recently, researchers further extends the scope of the affordance grounding problem, including extending it to scene understanding~\cite{qian2023understanding,kulal2023putting,chen2023affordance}, 3D models~\cite{yang2023grounding}, egocentric videos~\cite{mur2023multi}, hand pose generation~\cite{jian2023affordpose,ye2023affordance}, or associating it with human parts~\cite{luo2023leverage}.
We use AGD20K as our primary benchmark, and compare our approach with state-of-the-art methods~\cite{qian2023understanding,nagarajan2019grounded,luo2022grounded,hadjivelichkov2023one,li2023locate}.

\mypar{Incorporating 3D Information for Vision Tasks.}
3D information has been shown to be critical in certain vision and robotics tasks. 
For example, Zhou \etal~\cite{zhou2019does} found that visual navigation in mobile sensorimotor systems can benefit from 3D input.
Kerr \etal~\cite{kerr2022evo} found the NeRF-rendered depth map can help grasping in robotics.
Similarly, grounding affordance could benefit from 3D information as well, as 3D shapes of objects and their parts hold a lot of hints on their utility and the proper ways to interact with them. 
While 3D information is not usually available for an arbitrary image, fortunately, researchers have built a series of robust approaches to estimate the 3D of an image, ranging from surface normal estimation~\cite{wang2015designing,eigen2015predicting}, depth estimation~\cite{chen2019learning,li2018learning,ranftl2021vision,yin2021learning,yin2019enforcing}, to 3D reconstruction~\cite{Gkioxari2019,liu2019planercnn,nie2020total3dunderstanding} and few-image NeRF~\cite{liu2023zero,qian2021recognizing,cao2022fwd,yu2021pixelnerf}.
In our paper, we mainly use DPT~\cite{ranftl2021vision} to generate pseudo depth maps to help VLMs to build 3D understanding.

\mypar{Robotics Manipulation.}
Manipulation of in-the-wild objects is an importatnt but challenging task in robotics due to the difficulty of data collection.
Researchers have developed many methods for different objects in different scenes, such as tabletop objects~\cite{james2020rlbench,goyal2023rvt,shridhar2023perceiver} and mobile manipulation~\cite{yenamandra2023homerobot}.
While manipulation is not our goal, learning affordance can be a solution for manipulation~\cite{bahl2023affordances,hsu2023ditto,wu2023learning}.

\section{Approach}
\label{sec:approach}

We now introduce our approach, \methodname, which takes a single image $I$ and an affordance text query $T$, and generates an affordance map $M$.
We use a template of ``What part of the \texttt{<object\_name>} should we interact with in order to \texttt{<action\_name>} it?'' as the text query $T$. 
We then train LLM to generate a special token \texttt{<mask\_token>} and use its hidden state to decode a dense affordance map $M$.
A brief overview is shown in Fig~\ref{fig:approach}.

\subsection{Overview}

\mypar{Large language model.}
We choose LLaVA-7B as our backbone multimodal large language model.
We refer the reader for
a fuller explanation in~\cite{liu2023llava}, but briefly, LLaVA contains an image encoder, a text tokenizer, and a large language model $\mathrm{LLM}$.
The image encoder is typically a CLIP pretrained ViT, with a linear layer to project the hidden dimension.
It encodes the image $I$ into image features $F_I$.
At the same time, the tokenizer encodes the text $T$ into text features $F_T$.
They are concatnated and feed into the language model. The $\mathrm{LLM}$ produces text output $A$ as:
\begin{equation}
    A = \mathrm{LLM}(F_I, F_T).
\end{equation}

\mypar{Predicting affordance.}
How do we perform affordance reasoning while leveraging the world knowledge embedded in LLM? We propose to treat affordance as an implicit text token predicted from the LLM, which could be further decoded into a 2D map. 
Specifically, we train the LLM to predict a special token \texttt{<mask\_token>}, the hidden state of which is first projected into a query embedding $q$ and then fed into a $\mathrm{Decoder}$ to generate a dense affordance map. 
$\mathrm{Decoder}$ shares a similar architecture as the ones in SAM~\cite{kirillov2023segany} and 3DOI~\cite{qian2023understanding}.
It takes in $q$ and image features $F_I$ to produce an affordance map $M$, i.e.,
\begin{equation}
    M = \mathrm{Decoder}(F_I, q).
    \label{eqn:decoder}
\end{equation}

\mypar{Pseudodepth as additional inputs.}
Besides images, the affordance reasoning task could benefit from 3D information (we will validate the benefits in Sec~\ref{sec:experiments}).
However, modern VLMs are typically only trained with text and 2D images~\cite{liu2023llava,li2023blip2}. Therefore, we also include a pseudo depth map as additional inputs to the large language model.
For each image, we use the state-of-the-art depth estimation model DPT~\cite{ranftl2021vision} to generate a pseudo depth map $D$.
We use the same image encoder to encode the depth map $D$ to produce depth features $F_D$, and concatenate it with image features. Our final model is thus 
\begin{equation}
    A, M = \mathrm{AffordanceLLM}(F_I, F_D, F_T).
\end{equation}

\mypar{Training objectives.}
Following the same practice as \cite{qian2023understanding}, we train the affordance map using a binary focal loss~\cite{lin2017focal}, and call it affordance loss $L_{\mathrm{aff}}$.
We set the weight of positive examples to be 0.95 and that of negative ones to be 0.05 to balance positives and
negatives, as there are more negatives than positives in ground truth affordance map.
We follow the standard cross entropy loss for the text output of language models.
Our final loss function is a linear combination of affordance loss and text loss, given by
\begin{equation}
    L = L_{\textrm{aff}} + \lambda \cdot L_{\textrm{text}}.
    \label{eq:objective}
\end{equation}

In practice, we set $\lambda = 0.01$ to balance two losses, as the affordance loss can be quite small due to the imbalance of positive and negative values. 

\subsection{Network Architecture}

Next, we discuss the network architecture, and the training details of our model.

\mypar{Image encoder.}
The standard LLaVA uses CLIP image encoder and a linear projection layer~\cite{xu2023masqclip,liu2023llava}.
In practice, we find that the CLIP image encoder has low resolution (224x224) and does not capture sufficient information about grounding.
Therefore, we use OWL-ViT~\cite{minderer2022simple} to replace the standard CLIP-ViT~\cite{radford2021clip}. 
OWL-ViT has an input resolution of 768x768, which is significantly higher than CLIP.
At the same time, OWL-ViT is pretrained to extract features that include precise location information of objects.
As we will empirically show in experiments, using OWL-ViT is significantly better than CLIP. However, we note that our approach is general, and will benefit from any future improvements in pretrained VLM backbones.

\mypar{Projection.}
Another problem of using OWL-ViT is about its embedding space. With a much higher input resolution, OWL-ViT produces 576 tokens with a hidden dimension of 768 for each image.
In comparison, CLIP only produces 256 tokens for each image.
Projecting each individual token into the hidden dimension of LLM (4096) consumes a lot of GPU memory. 
In practice, we project each token of OWL-ViT to 1024, and concatenating four neighboring tokens into a single token.

\mypar{Language model.}
We follow LLaVA~\cite{liu2023llava} and LLama~\cite{touvron2023llama} to use the standard text tokenizer to encode our text query.
We use LLama-7B~\cite{touvron2023llama} as the large language model.

\mypar{Affordance decoder.}
We aim to keep a lightweight decoder, as it has been proved to produce good segmentation masks and affordance maps~\cite{kirillov2023segany,qian2023understanding,carion2020end,cheng2021per}.
However, we find the vanilla mask decoder is too lightweight in our case and does not produce high-resolution affordance map.
Therefore, we add an additional transposed convoluation layer to increase its output resolution.

\mypar{Implementation.}
We implement our model using PyTorch and HuggingFace.
We initialize our model with LLama-7B pretrained weights.
Following LLaVA~\cite{liu2023llava}, we freeze the image encoder, pretrain the image projection layer to align OWL-ViT and LLama features, and then use GPT instructions to tune the language model.
Finally, we add the mask encoder~\cite{kirillov2023segany,qian2023understanding} and tune the whole model on AGD20K~\cite{luo2022learning}, which has annotations of object affordance.
We use eight NVIDIA A100 (40GB) to train our model, with Fully Sharded Data Parallel. 
We use a batch size of 4 and set the learning rate as 2e-5.
\section{Experiments}
\label{sec:experiments}

In experiments, we aim to systematically evaluate the performance of our approach.
In particular, we are interested in answering these questions:
(1) How well does it generalize, compared with state-of-the-art methods?
(2) How does each design choice contribute to the final performance, including prompts, visual encoders, and depth?

\subsection{Experimental Setup}

\begin{figure*}[t]
  \centering
   \includegraphics[width=1\linewidth]{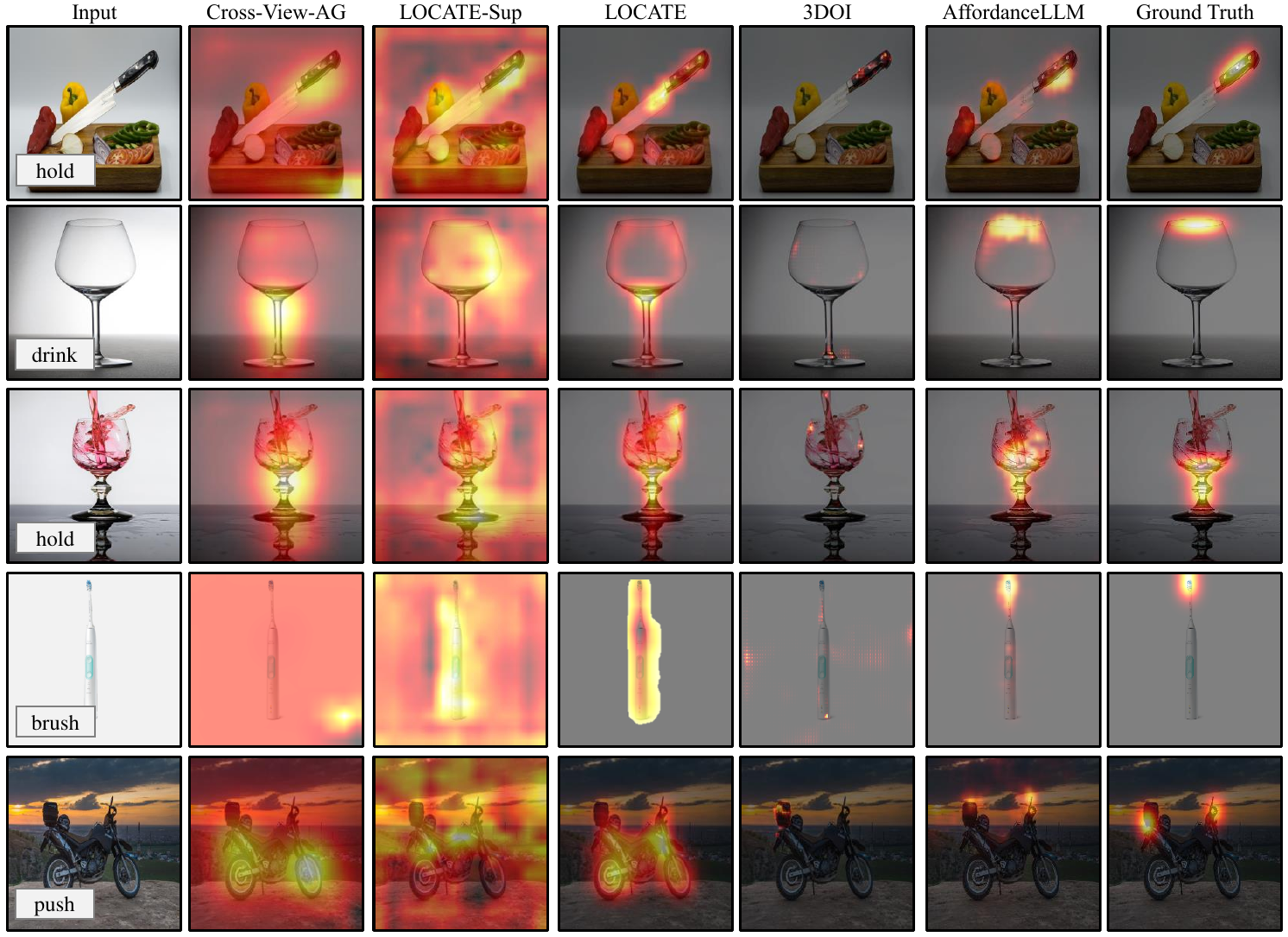}
    \vspace{-2em}
   \caption{Qualitative results on the test set of the hard split. 
   LOCATE-Sup fails to learn a reasonable affordance map due to limited training data.
LOCATE~\cite{li2023locate} typically predicts an affordance map which covers the whole object.
3DOI~\cite{qian2023understanding} focuses on a small area of the object.
Overall, our approach produces the best-quality affordance predictions.}
\vspace{-1em}
   \label{fig:results}
\end{figure*}

\mypar{Metrics.}
We evaluate primarily on AGD20K~\cite{luo2022learning} and follow its metrics to evaluate our model, which is KLD, SIM and NSS~\cite{nagarajan2019grounded,luo2022learning,qian2023understanding,li2023locate}.
For KLD, the lower the better. And for SIM and NSS, the higher, the better.
A full explanation is avilable in the supplemental.

\mypar{Baselines.}
We compare our approach against state-of-the-art baselines.
In general, affordance grounding methods belong to two categories: weakly supervised and fully supervised methods.
We report performance of both categories.

\noindent
\emph{(Weakly supervised methods):} 
They do not train on explicit labels of the affordance map.
Instead, they are trained on a human demonstration of the same object. These approaches include InteractionHotspots~\cite{nagarajan2019grounded}, Cross-View-AG~\cite{luo2022learning}, Cross-View-AG+~\cite{luo2022grounded}, AffCorrs~\cite{hadjivelichkov2023one}, and LOCATE~\cite{li2023locate}.
Among them, LOCATE is the most recent model and has the best results on AGD20K.
We use the reported number in LOCATE for the easy split and retrain them for the hard split.
Among them, we cannot run AffCorrs, as it focuses on one-shot affordance learning.
The reported model on the easy split is adapted by \cite{li2023locate} and not publicly available.
We also do not run InteractionHotspots because the pretrained model only supports 7 actions.
The reported model is retrained by \cite{luo2022learning} but lacks sufficient implementation details to be reproduced.
Therefore, we retrain Cross-View-AG~\cite{luo2022learning}, Cross-View-AG+~\cite{luo2022grounded}, and LOCATE~\cite{li2023locate} on the hard split.
We maintain the object/action split, but allow them to use more images for weak supervision.
Therefore, they has 11,889 images for training.

\noindent
\emph{(Fully supervised methods):} Affordance map can also be learned from explicit labels, and we call it supervised methods.
This includes 3DOI~\cite{qian2023understanding}
and ours.
We also adapt LOCATE to a fully supervised version for fair comparison.

\noindent
$\bullet$ \emph{3DOI~\cite{qian2023understanding}:}
3DOI is a SAM-based model~\cite{kirillov2023segany}, which takes a single image and a query point and predicts the segmentation mask and affordance map.
Therefore, We randomly sample a pixel with score $>0.9$ as the query point from the affordance map. 
We use the 3DOI pretrained model, which has never seen any images in AGD20K, including the training set.

\noindent
$\bullet$ \emph{LOCATE-Sup~\cite{li2023locate}:}
To ensure fair comparison, we also adopt LOCATE and train it using the same binary focal loss as our model. We call it LOCATE-Sup.
LOCATE uses a Dino-ViT as its visual encoder~\cite{caron2021emerging}.
To eliminate the effect of different pretrained visual encoders, we also report the performance of LOCATE-Sup-OWL, which uses the exact same pretrained visual encoder as ours.

\subsection{Dataset}
\label{sec:dataset}
We follow LOCATE~\cite{li2023locate} to evaluate primarily on AGD20K~\cite{luo2022learning}, as it is the only large-scale dataset for affordance with action and object labels.
Since our approach is not weakly supervised and requires dense annotations, we only use AGD20K images with dense annotations.

\begin{table}
   \centering
   \caption{Difficulty score of different splits. The lower the score, the more similar are the object categories in the train and test set. 
   }
   \label{tab:splits}
   \scalebox{0.9}{
   \begin{tabular}{lcccc}
      \toprule
      Splits & Same & Easy & Hard & Random \\
      \midrule
      Difficulty Score $\uparrow$ & 0.000 & 0.356 & 0.412 & 0.491\\
      \bottomrule
   \end{tabular}
   } %
\end{table}

\begin{table}
   \centering
   \caption{\textbf{Quantitative results on the \emph{Easy} split of AGD20K~\cite{luo2022learning}.} InteractionHotspots, Cross-View-AG(+), AffCorrs and LOCATE are trained on AGD20K images with weak supervision (13,323 images). LOCATE-Sup and LOCATE-Sup-OWL, and \methodname\ are trained on AGD20K images with dense annotation (1,135 images). 3DOI is trained on their own dataset with dense annotation (10,000 images)~\cite{qian2023understanding}. \methodname\ is comparable to LOCATE~\cite{li2023locate} on the easy split, where test objects have similar counterparts in the training set. The \textbf{best} and \underline{second-best} results are highlighted in bold and underlined, respectively.}
   \label{tab:main_easy}
   \vspace{-0.8em}
   \scalebox{0.95}{
   \begin{tabular}{llccc}
      \toprule
      Methods & KLD $\downarrow$ & SIM $\uparrow$ & NSS $\uparrow$ \\
      \midrule
    InteractionHotspots~\cite{nagarajan2019grounded}  & 1.994 & 0.237 & 0.577  \\
      Cross-View-AG~\cite{luo2022learning}  & 1.787 & 0.285 & 0.829    \\
      Cross-View-AG+~\cite{luo2022grounded}  & 1.765 & 0.279 & 0.882   \\
      AffCorrs~\cite{hadjivelichkov2023one}  & 1.618 & 0.348 & 1.021 \\
    LOCATE~\cite{li2023locate}  & \textbf{1.405} & \underline{0.372} & \textbf{1.157} \\
    LOCATE-Sup~\cite{li2023locate} &  1.907 & 0.236& 0.641  \\
    LOCATE-Sup-OWL~\cite{li2023locate,minderer2022simple}   & 1.927 & 0.234& 0.624   \\
      3DOI~\cite{qian2023understanding} & 3.565 & 0.227 & 0.657 \\
      \textbf{\methodname\ (Ours)} & \underline{1.463} & \textbf{0.377} & \underline{1.070} \\
      \bottomrule
   \end{tabular}
   } %
   \vspace{-1em}
\end{table}

In this paper, we primarily evaluate the ability of an affordance model to generalize to unseen object categories, and thus evaluate on the \textit{Unseen} split of the AGD20K benchmark. 
This split ensures that there is no overlap between the object categories in the train and test set.

\begin{figure*}
    \centering
    \includegraphics[width=\linewidth]{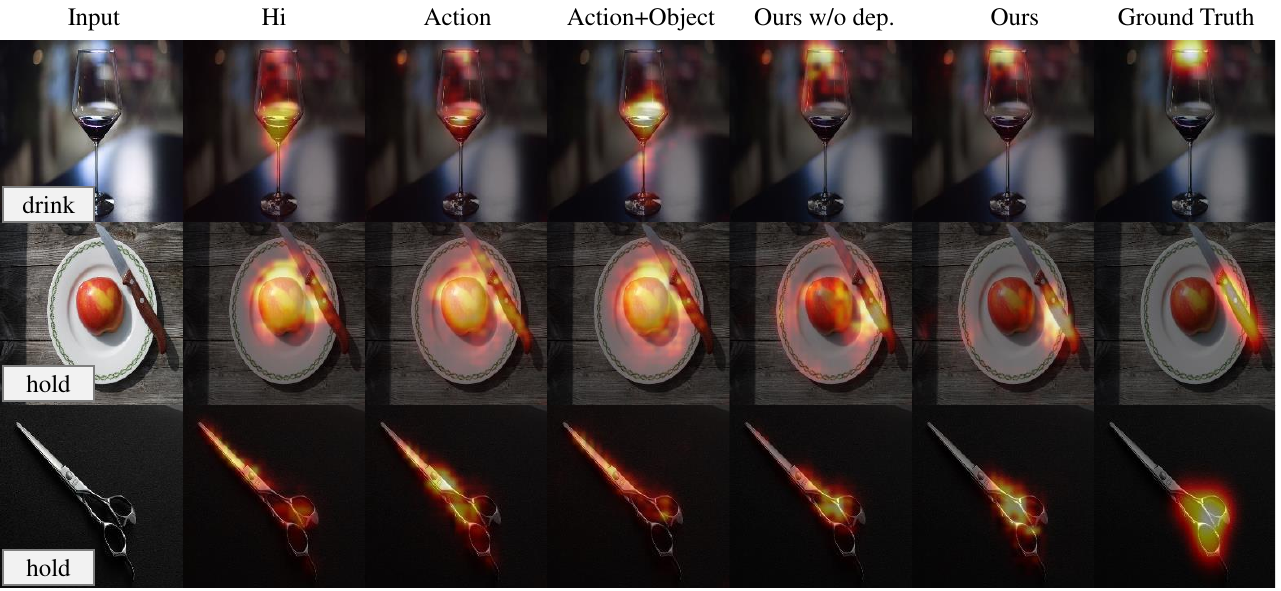}
    \vspace{-2em}
    \caption{Ablation of different text prompts and depth. 
    Ours w/o depth is our approach without pseudodepth as additional inputs. Ours is our full approach. We find constructing the correct text prompt typically helps our model to focus on the correct area. We believe it is because the correct text prompt would activate the world knowledge related to affordance embedded in the VLM.}
    \label{fig:ablation}
    \vspace{-1em}
\end{figure*}

However, the \textit{Unseen} split has one major issue: there are still a lot of similarities between the objects in the train and test set.
Objects in the test set typically have similar counterparts in the training set, leaving models room for memorizing.
For example, ``skis'' in the test set maps to ``snowboards'' and ``skateboards'' in training set,
``basketball'' maps to ``baseball'', ``knife'' maps to ``fork'', etc.
To make the benchmark more reflective of the generalization ability of a model, we provide a more challenging split. We thus have the following two splits.

\begin{table}
   \centering
   \caption{\textbf{Quantitative results on the \emph{Hard} split of AGD20K~\cite{luo2022learning}.} Cross-View-AG(+) and LOCATE are trained on AGD20K images with weak supervision (11,889 images). LOCATE-Sup and LOCATE-Sup-OWL, and \methodname\ are trained on AGD20K images with dense annotation (868 images). 3DOI is trained on their own dataset with dense annotation (10,000 images)~\cite{qian2023understanding}. On the hard split, \methodname\ outperforms all baselines by a large margin, which demonstrates the superior generalization ability of our model.
   We do not run InteractionHotspots~\cite{nagarajan2019grounded} and AffCorrs~\cite{hadjivelichkov2023one}, as the reported model has ambiguous implementation details, or is not publicly available. The \textbf{best} and \underline{second-best} results are highlighted in bold and underlined, respectively.}
   \label{tab:main_hard}
   \vspace{-0.8em}
   \scalebox{0.95}{
   \begin{tabular}{llccc}
      \toprule
      Methods & KLD $\downarrow$ & SIM $\uparrow$ & NSS $\uparrow$ \\
      \midrule
      Cross-View-AG~\cite{luo2022learning}  & 2.092 & 0.209 & 0.138    \\
      Cross-View-AG+~\cite{luo2022grounded}  & 2.034 & 0.218 & 0.342   \\
    LOCATE~\cite{li2023locate}  & \underline{1.829} & \underline{0.282} & 0.276 \\
    LOCATE-Sup~\cite{li2023locate} &  2.003 & 0.224 & 0.435  \\
    LOCATE-Sup-OWL~\cite{li2023locate,minderer2022simple}   & 2.127& 0.206 & 0.314   \\
      3DOI~\cite{qian2023understanding} & 4.017 & 0.200 & \underline{0.549} \\
      \textbf{\methodname\ (Ours)} & \textbf{1.661} & \textbf{0.361} & \textbf{0.947} \\
      \bottomrule
   \end{tabular}
   } %
   \vspace{-1em}
\end{table}

\mypar{Easy split.}
This is the original \textit{Unseen} split of AGD20K.
We have 1135/540 images for train and test with dense annotations for the fully supervised setting, or 13,323/540 images for the weakly supervised setting. The test set remains the same for both settings.

\mypar{Hard split.}
We \textit{randomly} put 50\% AGD20K object classes into the training set and the remaining classes into the test set to simulate in-the-wild generalization (details in the supplemental).
The training and test object are not overlapping, and most objects in the test set do not have a similar counterpart in the training set, and is thus harder to generalize than the \textit{Easy} split.
We have 868/807 images for train and test with dense annotations for the fully supervised setting, and 11,889/807 images for weakly supervised setting.
The test set is the same for both settings.

\mypar{Measuring split difficulty.}
We propose a metric to quantify the generalization difficulty of a split. 
Intuitively, the difficulty to generalize to an object class in the test set is defined by how different it is from the classes in the training set, 
which could be measured by its semantic distance to the most similar class in the training set~\cite{Qian2020}.
The greater the distance, the harder it is to generalize to this test class.
Therefore, for each semantic class $c$ in the test set, we compute its distance $d$ to the most similar class in the training set.
We use the CLIP~\cite{radford2021learning} text encoder to obtain an embedding to represent each object class.
Assume train classes are $C_{\mathrm{train}}$ and test classes are $C_{\mathrm{test}}$,
the difficulty of this split is 
\begin{equation}
    D(C_{\mathrm{train}}, C_{\mathrm{test}}) = 1-\frac{1}{|C_{\mathrm{test}}|} \sum_{c \in C_{\mathrm{test}}} \max_{c' \in C_{\mathrm{train}}} d(c, c').
    \label{eqn:similarity}
\end{equation}

We compare the difficulty metric among four settings: (1) \textit{Same}: train and test share the same classes; (2) \textit{Easy} split; (3) \textit{Hard} split; (4) \textit{Random}: constructed by randomly even-splitting 50 object classes from LVIS~\cite{gupta2019lvis}, which serves as a lower bound.

We show the difficulty metrics in Tab~\ref{tab:splits}.
The \textit{Same} split has a similarity metric of 0.0, as all object classes in the test are present during training.
The \textit{Easy} split has a similarity metric of 0.356.
The \textit{Random} split has a score of 0.491.
The \textit{Hard} split has a higher score than \textit{Easy}, meaning that the difference between test and train is more significance in \textit{Hard} than in \textit{Easy}, and is thus harder to generalize.

\subsection{Results}

Figure~\ref{fig:results} shows qualitative results on the test set of the hard split. 
In this split, the objects in the test set bear little to none resemblance to the ones in the training set.
We compare our approach, \methodname, with a set of state-of-the-art baselines.
LOCATE~\cite{li2023locate} tends to predicts an affordance map that covers the entire object, indicating poor generalization performance.
3DOI~\cite{qian2023understanding} typically focuses on a small area of the object, and sometimes fails to ground the correct region.
LOCATE-Sup fails to predict reasonable affordance map, probably due to the small amount of training data.
Despite being trained on the same training set as LOCATE-Sup, our approach is able to produce the best affordance map among all methods, showcasing superior generalization capability.

We further compare our model with baselines quantitatively and the results are summarized in Tab.~\ref{tab:main_easy},~\ref{tab:main_hard}.
On the \textit{Hard} split, where the test set objects differ semantically from the training set, our method outperforms all baselines significantly. This improvement can be attributed to the extensive world knowledge and understanding embedded within the large language model.
On the \textit{Easy} split, our model is comparable to LOCATE~\cite{li2023locate} and outperform all other baselines. 
We hypothesize that the advantage of our approach is less pronounced when the test and train objects exhibit similarity, as the generalization capability becomes less critical. 
It is also worth noting that unlike LOCATE which is weakly supervised on 10k+ images, our model was fully supervised on some 1k images with dense annotations, which renders a more meaningful comparison with LOCATE-Sup that is trained on the same data. 
Our method significantly outperforms LOCATE-Sup on both splits, indicating the effectiveness of our approach.

\begin{table}
   \centering
   \caption{Ablation on the hard split. We validate the importance of text prompts, image encoders and pseudo depth to performance. 
   }
   \vspace{-0.8em}
   \label{tab:ablation}
   \scalebox{0.82}{
   \begin{tabular}{cccccc}
      \toprule
      Depth & Text Prompt & Img Encoder & KLD $\downarrow$ & SIM $\uparrow$ & NSS $\uparrow$\\
      \midrule
       Yes & Full & OWL-ViT & \textbf{1.661} & \textbf{0.361} & \textbf{0.947}  \\
       - & Full & OWL-ViT & 1.713 & 0.352 & 0.881  \\
       - & Full & CLIP-ViT & 1.759 & 0.286 & 0.776  \\
       - & Object, Action & OWL-ViT & 1.769 & 0.329 & 0.827  \\
       - & Action & OWL-ViT & 1.843 & 0.336 & 0.815  \\
       - & Hi & OWL-ViT & 1.836 & 0.325 & 0.793  \\
      \bottomrule
   \end{tabular}
   } %
   \vspace{-1em}
\end{table}

\subsection{Ablation}

We conduct a few ablation studies to understand how different components of the model contribute to the final performance.
We test different text prompts, different image encoders, and the effect of pseudo depth as inputs.
The results are summarized in Tab~\ref{tab:ablation} and Fig~\ref{fig:ablation}.

\mypar{Text prompts.}
Prompt tuning is known to have major effects on large VLMs.
We test four different text prompts to understand the effect of text content on model performance:

\noindent
$\bullet$ \emph{Hi:} We use ``Hi'' as our text prompt.

\noindent
$\bullet$ \emph{Action:} We use the action (e.g. ``hold'') as the prompt.

\noindent
$\bullet$ \emph{Object + Action:} We use the object name and action label as our text prompt, for example ``hold, knife''.

\noindent
$\bullet$ \emph{Full:} We use a complete question as the text prompt --- ``What part of the motorcycle should we interact with in order to push it?''.

We notice that the \emph{Full} prompt yields a higher performance compared with other simple text prompts.
It demonstrates that specific question prompt is helpful for extracting the knowledge from pretrained large language models.

\begin{figure}[t]
  \centering
   \includegraphics[width=1\linewidth]{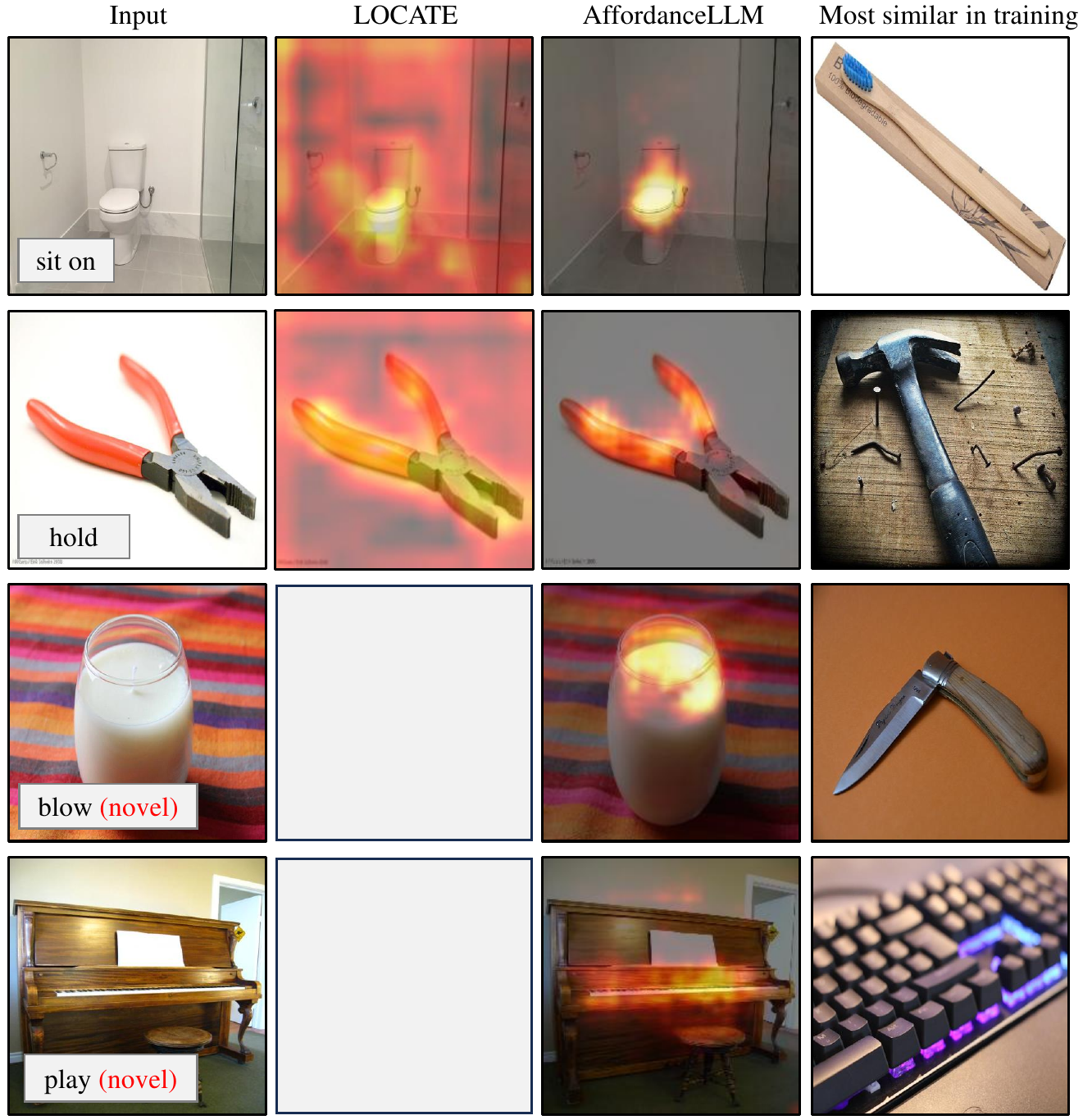}
   \vspace{-2em}
   \caption{Generalization results on random Internet images. We show the most similar objects in the training set to demonstrate how different the objects are from the ones in the training set. \textbf{(Row 1, 2):} \methodname\  generalizes to novel objects from random Internet images, while LOCATE~\cite{li2023locate} fails. \textbf{(Row 3, 4):} \methodname\ generalizes to novel actions plus novel objects. LOCATE cannot infer novel actions thus we left it blank.}
   \vspace{-1em}
   \label{fig:generalization}
\end{figure}

\mypar{Vision encoders.}
Although LLaVA~\cite{liu2023llava} uses CLIP-ViT, it may not be the optimal vision encoder for our affordance grounding task --- 
CLIP-ViT is trained with an objective to align text-image pairs and is not explicitly optimized to perform localization, and therefore has limited visual grounding capability.
We therefore switch to OWL-ViT~\cite{minderer2022simple}, which is trained on detection datasets with 2M images, and achieves state-of-the-art open vocabulary detection performance. 
As shown in Tab~\ref{tab:ablation}, using OWL-ViT as vision backbone far excels using CLIP-ViT. It indicates the importance of grounding capability of visual backbone.

\subsection{Pseudodepth as Inputs}
Our model is trained with pseudo depth map produced by DPT~\cite{ranftl2021vision}.
To verify whether the additional depth inputs are effective, we compare the model trained with and without estimated depth (Tab~\ref{tab:ablation} and Fig~\ref{fig:ablation}).
With depth, our model can predict better affordance map, demonstrating the importance of 3D information in affordance reasoning.

\begin{figure}[t]
  \centering
   \includegraphics[width=1\linewidth]{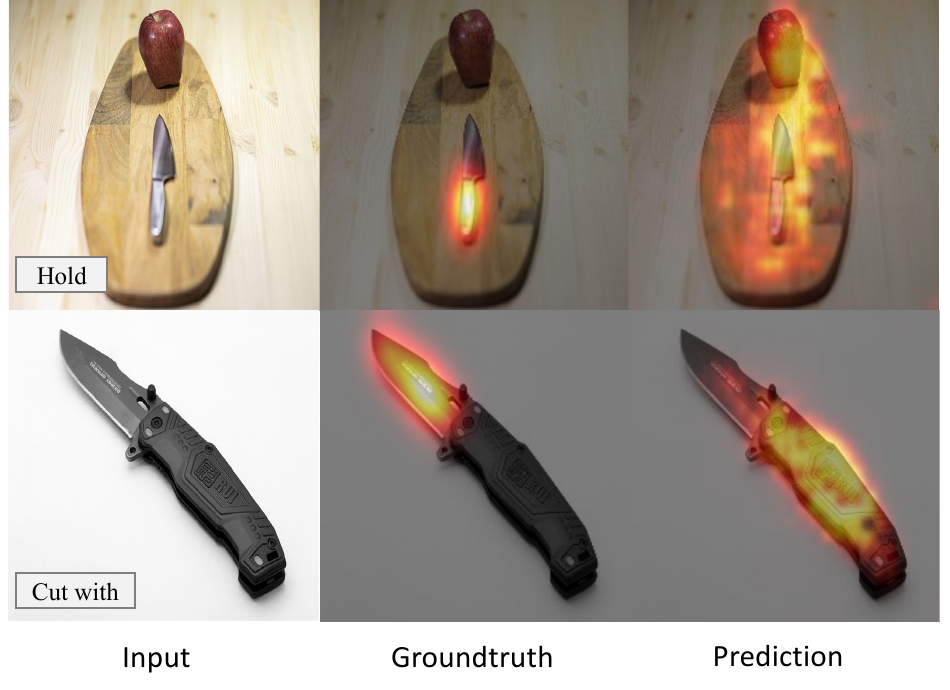}
    \vspace{-2em}
   \caption{Failure examples. \textbf{(Row 1:)} \methodname\ sometimes fails due to multiple objects present in the scene and it fails to refer to the correct object. \textbf{(Row 2:)} \methodname\ thinks humans should hold the handle to cut something using the knife, while AGD20K annotators think ``cut with'' should refer to the blade.}
   \vspace{-1em}
   \label{fig:fail_cases}
\end{figure}

\subsection{Generalization to Internet Images}

We further test the generalization of our model on random Internet images in Fig.~\ref{fig:generalization}.
All test objects are novel. To showcase how different these objects are from the ones in train set, for each test object, we retrieve the most similar object in the train set using the metric defined in Eq~\ref{eqn:similarity}. 
As we can see, these objects are not similar to any objects in the train set.
We go even further to test if our approach is able to generalize to novel actions in addition to novel objects, such as ``blow'' and ``play''.
Generalization to novel actions is even more challenging, requiring open vocabulary understanding of actions, which is beyond the capability of LOCATE~\cite{li2023locate}.
Despite these challenges, our approach not only produces very reasonable affordance maps for novel objects, but is also able to handle novel actions plus novel objects, once again demonstrating the extraordinary capability to generalize.

\subsection{Failure Examples}

Finally, we show our failure examples in Fig~\ref{fig:fail_cases}. 
First, we find \methodname\ fails on some ambiguous questions.
For example, in AGD20K, ``cut with'' refers to the blade of a knife.
However, \methodname\ thinks humans should hold the handle of the knife to cut anything.
Second, when there are multiple objects in the image, it sometimes cannot refer to the correct object.

\section{Conclusion}

We have presented \methodname, a novel approach which can ground affordance for in-the-wild objects unseen during training.
By tapping into the world knowledge embedded in a Vision Language Model, our proposed approach generalizes much better to in-the-wild objects, compared with state-of-the-art affordance grounding models.

Our approach can have positive impacts by helping build intelligence robots which can manipulate in-the-wild objects.
On the other hand, it can be misused to cause physical damage or harm if applied in an adversarial manner.

{
    \small
    \bibliographystyle{ieeenat_fullname}
    \bibliography{local}
}

\clearpage
\appendix

\section{Metrics}

In the section, we explain the metrics (KLD, SIM, and NSS) to evaluate our model. 

\vspace{0.75em}
\noindent
$\bullet$ \textbf{K}ullback-\textbf{L}eibler \textbf{D}ivergence (KLD) measures distribution difference between the predicted affordance map ($M$) and the ground truth ($M'$), which is 
\begin{equation}
\small
   \mathrm{KLD}\left ( M,M' \right )=\sum_{i}M'_{i}\log\left ( \epsilon + \frac{M'_{i}}{\epsilon+M_{i}} \right ), \label{eq:no20}
\end{equation}

\noindent
$\bullet$ \textbf{Sim}iliary (SIM) is also called histogram intersection, which measures the intersection between the predicted affordance map ($M$) and the ground truth ($M'$). The final range is from 0 to 1. It is given by

\begin{equation}
   \mathrm{SIM}\left ( M, M' \right )=\sum_{i}\min\left ( M_{i},M'_{i}\right ),\\
\end{equation}
where  $\sum_{i}M_{i}=\sum_{i}M'_{i}=1$.

\noindent
$\bullet$ \textbf{N}ormalized \textbf{S}canpath \textbf{S}aliency (\textbf{NSS}) measures the correspondence between the prediction map ($M$) and the ground truth ($M'$). It is given by

\begin{equation}
\small
   \mathrm{NSS}\left ( M,M' \right )=\frac{1}{N}\sum_{i}\hat{M}\times M'_{i}, \label{eq:no22}
\end{equation}

where $N=\sum_{i}M'_{i}$, $\hat{M}=\frac{M-\mu\left ( M \right )}{\sigma\left ( M \right )}$. $\mu\left ( M \right )$ and $\sigma\left ( M \right )$ are the mean and standard deviation, respectively.

\section{Details of the Data Splits}

In the easy split, we follow the object split of the orginal AGD20K \emph{Unseen} setting~\cite{luo2022learning}. 
The easy split has 33 object classes for training and 14 for testing.
We have 1135/540 images for train and test with dense annotations for the fully supervised setting, or 13,323/540 images for the weakly supervised setting.

\begin{itemize}
    \item Train classes: scissors, badminton racket, surfboard, frisbee, hot dog, tennis racket, hammer, microwave, oven, punching bag, carrot, snowboard, book, suitcase, skateboard, wine glass, keyboard, javelin, motorcycle, discus, bench, toothbrush, bottle, cell phone, chair, orange, rugby ball, couch, baseball, fork, bowl, apple, baseball bat.

    \item Test classes: camera, bed, bicycle, golf clubs, soccer ball, cup, laptop, banana, skis, knife, axe, broccoli, basketball, refrigerator.
\end{itemize}

In the hard split, we randomly put around 50\% AGD20K object classes
into the training set and the remaining classes into the test
set to simulate in-the-wild generalization.
The hard split has 28 object classes for training and 22 for testing.
We have 868/807 images for train and test with dense annotations for the fully supervised setting, and 11,889/807 images for the weakly supervised setting.

\begin{itemize}
    \item  Training objects include carrot, cup, bowl, discus, book, camera, golf clubs, bottle, broccoli, binoculars, drum, baseball, apple, frisbee, cell phone, baseball bat, couch, hammer, bicycle, bench, fork, badminton racket, banana, hot dog, axe, bed, chair, basketball.
    \item Test objects include soccer ball, laptop, punching bag, oven, suitcase, javelin, wine glass, motorcycle, scissors, snowboard, keyboard, rugby ball, orange, surfboard, knife, skateboard, pen, microwave, skis, tennis racket, refrigerator, toothbrush.
\end{itemize}

\end{document}